\documentclass[sigconf,natbib=true,anonymous=false]{acmart}
\renewcommand\footnotetextcopyrightpermission[1]{} 
\usepackage{multirow}
\usepackage{booktabs} 
\AtBeginDocument{%
  \providecommand\BibTeX{{%
    \normalfont B\kern-0.5em{\scshape i\kern-0.25em b}\kern-0.8em\TeX}}}

\begin{document}

\title{Zero-shot Entity and Tweet Characterization with \\ Designed Conditional Prompts and Contexts}

\author{Sharath Srivatsa}
\email{sharath.srivatsa@iiitb.ac.in}
\affiliation{%
  \institution{International Institute of Information Technology, Bangalore \country{}}
}
\author{Tushar Mohan}
\email{tushar19393@iiitd.ac.in}
\affiliation{%
  \institution{Indraprastha Institute of Information Technology, Delhi \country{}}
}
\author{Kumari Neha}
\email{nehak@iiitd.ac.in}
\affiliation{%
  \institution{Indraprastha Institute of Information Technology, Delhi \country{}}
}
\author{Nishchay Malakar}
\email{nishchay.sr@iiitb.ac.in}
\affiliation{%
  \institution{International Institute of Information Technology, Bangalore \country{}}
}
\author{Ponnurangam Kumaraguru}
\email{pk.guru@iiit.ac.in}
\affiliation{%
  \institution{International Institute of Information Technology, Hyderabad \country{}}
}
\author{Srinath Srinivasa}
\email{sri@iiitb.ac.in}
\affiliation{%
  \institution{International Institute of Information Technology, Bangalore \country{}}
}

\begin{abstract}
Online news and social media have been the de facto mediums to disseminate information globally from the beginning of the last decade. However, bias in content and purpose of intentions are not regulated, and managing bias is the responsibility of content consumers. In this regard, understanding the stances and biases of news sources towards specific entities becomes important. To address this problem, we use pretrained language models, which have been shown to bring about good results with no task-specific training or few-shot training. In this work, we approach the problem of characterizing Named Entities and Tweets as an \textit{open-ended text classification} and \textit{open-ended fact probing} problem. We evaluate the zero-shot language model capabilities of \textit{Generative Pretrained Transformer 2} (GPT-2) to characterize Entities and Tweets subjectively with human psychology-inspired and logical conditional prefixes and contexts. First, we fine-tune the GPT-2 model on a sufficiently large news corpus and evaluate subjective characterization of popular entities in the corpus by priming with prefixes. Second, we fine-tune GPT-2 with a Tweets corpus from a few popular hashtags and evaluate characterizing tweets by priming the language model with prefixes, questions, and contextual synopsis prompts. Entity characterization results were positive across measures and human evaluation. 
\end{abstract}

\maketitle
\pagestyle{plain} 
\section{Introduction}
Online News and Social Media content have maximum reach and readers. The content often influences the reader, often changing worldviews and consequent decisions. Each news source comes with its own perspective and biases, and it is detrimental to assume that readers would discern the import and authenticity of the content. There is an increasing need to computationally characterize the diverse perspectives on content from online media and social media.

In the recent past, researchers have devised various approaches to tag wide-ranging, malicious and biased content and alert the readers~\cite{NewFaceb3:online, Facebook61:online, Building9:online}. Major social media organizations have implemented checks and transparency measures to spot affective content and provide metainformation of the content for alert users to report. Widely accepted empirical ways of detecting bias and misinformation content include validating against some ground truth, checking the writing style, inherent nature of information, and credibility of the source. Adversarial biased contents evolve, making their classification with hard-labels impractical, requiring the detection systems to adapt to the changes. Given the evolving nature of the adversarial content, we focus on the subjective approach to solve content characterization. In the near future, massively trained multi-task learner language models could play a crucial role in building reliable and resilient Internet content validation systems. 

We define \textit{Entity characterization} as saying something about a person pertinent in a timeline or a context and \textit{Tweet characterization} as saying something about intention or purpose of the Tweet. We construct human-psychology-inspired design prompts for entity characterization. The prompts constitute common constructs humans use in spoken and written format about a person. On the similar lines, we construct contextual synopsis, questions and templates which a human-being uses or is aware-of to understand an informative sentence like a Tweet.

In this work, we evaluate approaches for zero-shot characterization of Entities and Tweets. On news corpora, we evaluate entity characterization with an approach named \textit{``Designed Conditional Prefix-prompts.''} News articles have rich descriptive content on popular entities like persons, places, etc. and are a good source for language models to learn about entities. We finetune GPT-2 on a large corpus of news articles from seven different news media houses. We acquire news article URLs from \textit{Global Database of Events, Language, and Tone} (GDELT Project\footnote{GDELT: \url{https://www.gdeltproject.org/}}) and scraped  web-page content. Next, we compile a list of popular entities across news media houses for our experiments. We used common English language phrases to describe an entity as prefix-prompts. Similarly, we also experimented with entity characterization of entities appearing in tweets corpus. We hypothesize that \textit{Pretrained Language Models} (PLMs) finetuned with a large corpus of descriptive knowledge about entities can be leveraged as proxy experts to characterize entities subjectively with respect to the corpus.

We evaluate Tweet characterization on a corpora of Tweets from popular hashtags with an approach named \textit{``Designed Contextual Questions and Templates.''} We collate a Tweets corpora with varied Emotions, Emotionally Manipulative Language (EML), Prejudice, Dogmatism, and Call-to-Action which collectively represent biased perspectives, to fine-tune the PLM and evaluate zero-shot subjective Tweet characterization. We hypothesize that PLMs finetuned with a particular set of concepts can be leveraged as an expert to deduce such concepts.
 
\section{Related Work}

Information shared on social media might contain various forms of manipulation to stir targeted emotion in the reader. Researchers have focused on using crowdsourced methods for subjective judgments like identifying emotionally manipulative language (EML) \cite{huffaker2020crowdsourced}, propaganda \cite{barron2019proppy}, bias \cite{spinde2020enabling}, or prejudice \cite{wei2020examining}.
The state-of-the-art approach to identify emotionally manipulative language is based on a crowdsourced method that neutralizes the effect of intrinsically manipulative language by measuring EML through comparison. Apart from the crowdsourced approach, lexicon-based approaches have been used to identify emotional content in the text snippet. The use of classification-based methods backed by crowdsourced data is limited in their ability to generalize knowledge past what exists in training data, making them vulnerable to new patterns and references not explicitly trained for.

Recent few-shot approaches in Text Classification with language models have promising results. \citeauthor{schick2021exploiting} \cite{schick2021exploiting}, show text classification by training with a masked (label to be predicted) token pattern or a \textit{cloze question} pattern. Classification training is started with a small set of data, and with a semi-supervised approach, training data is increased with soft labels. An ensemble of language models is used during each step of the semi-supervised training. The final classifier is trained on sufficiently large soft-labeled data resulting from iterations of semi-supervised training. Authors have experimented with different numbers on initial training data; key observation from this work is that good results are observed in zero-shot or without training data, which supports our approach to experiment with zero-shot.

In \citeauthor{schick2021s} \cite{schick2021s}, on SuperGLUE  language tasks, pattern exploiting training is applied with a small language model ALBERT. The results are as good as GPT3, a large language model. Few SuperGlue tasks, when converted to cloze questions, require the prediction of multiple masked tokens, and an approach for this is detailed. The observation is that small language models lead to good results by leveraging cloze questions in text classification. We experimented with designed Contextual Synopsis, Boolean Questions, and Open-ended Questions in Tweet characterization in a zero-shot setting.

In \citeauthor{schick2020automatically} \cite{schick2020automatically}, labels for text classification are mined from the language model for text classification. The auto-generated labels are near-synonyms of labels in the training data. The observation is that the language models have sufficient knowledge to generate tokens in a context. We have experimented with subjective entity characterization in a zero-shot setting.

In \citeauthor{hambardzumyan-etal-2021-warp} \cite{hambardzumyan-etal-2021-warp}, text classification is trained by learning an embedding from continuous space, which, when precedes the masked token, a classification label, instructs the language model to generate a high probable and most appropriate token in the location of the mask. The critical observation is that the underlying pretrained language model has sufficient prompt knowledge, and inferring appropriate embedding to solve classification tasks is possible. 

In \citeauthor{gao-etal-2021-making} \cite{gao-etal-2021-making}, the authors state that classification tasks with zero-shot have limitations and propose a suite of few-shot approaches for classification tasks—Fine-tuning PLM with prompt-based learning and demonstrations and automatic prompt, label, and template generations. We evaluated if PLMs finetuned with domain data-set have sufficient knowledge to characterize entities and tweets.

Common-sense knowledge mining with zero-shot learning is another area that has given promising results recently. In \citeauthor{davison-etal-2019-commonsense} \cite{davison-etal-2019-commonsense}, a uni-directional language model (GPT-2) is first used to convert information triplets to a valid and highly probable sentence, and then a bi-directional language model (BERT-large) is used to score the validity of the given fact. This is done by calculating the weighted point-wise mutual information (PMI) for a given relation. The testing was done on PLMs or general models rather than finetuned ones to avoid biasing towards a given database. The results suggest that the unsupervised techniques outperform the currently available supervised approaches. This backs our approach of not training the model to downstream tasks and using ju language model finetuned on contextual data to extract knowledge-based a given prompt.

This leads to probing facts from pretrained language models (PLMs). The language models have numerous advantages over knowledge bases as they do not require any schema engineering, are easy to extend over more data, and are robust for unsupervised training. In \citeauthor{petroni2019language} \cite{petroni2019language}, the authors of the paper proved the strong ability of the PLMs to recall factual knowledge without any task-specific fine-tuning and their usefulness as open-domain QA systems. In \citeauthor{kassner2020negated} \cite{kassner2020negated}, authors use LAnguage Model Analysis \textit{(LAMA)}  \citeauthor{petroni2019language} \cite{petroni2019language} benchmark datasets and show that negation of an affirmative sentence has no effect on the models prediction of the masked word. Moreover, prepending a given sentence with any misprime(s) mislead the model to predict wrong tokens. The authors of \citeauthor{jiang2020x} \cite{jiang2020x} extend the LAMA benchmark dataset to extract information in various different languages. They custom-made templates to be converted into sentences in different languages and prompted the model to fill in the masked word(s). In \citeauthor{DBLP:conf/acl/KumarT21} \cite{DBLP:conf/acl/KumarT21}, the employed few-shot learning with as few as 25 annotated examples. The examples were permuted with their proposed modification in the genetic algorithm. Then the sequence was used as a prompt to the model along with a new sample, which was to be predicted for the intended task. 

The authors of \citeauthor{bragg2021flex} \cite{bragg2021flex} introduce a prompt-based model for a few short learning where the key idea was to pose prompts in the form of multiple-choice questions (MCQs). This backs our approach to characterize the tweets with templates of context synopsis and text to be classified followed by a multiple-choice question.

Nishida et al. \cite{nishida2020unsupervised} introduced a sequential finetuned BERT model for reading comprehension tasks. They accommodated an unsupervised learning method to make the domain adaptation of the BERT model on the target source, after which the model was further finetuned to source domain reading comprehension (RC) dataset. This way, the BERT was able to take care of RC tasks on a domain different from the one it was pretrained on. This backs our approach to use domain adaptation of Pretrained Language Models (PLMs) as a tool for getting domain-specific answers on prompting with task described inputs.

Domain-Adaptive Pretraining (DAPT) of the PLMs has proved to greatly improve the model's performance on tasks from the target domain. A PLM was separately fine tuned on four different domain datasets, namely biomedical, computer science, news and reviews, and was then tested on corresponding two tasks of each domain. For each task, the domain adapted model performed better \cite{gururangan2020don}.

These approaches back our way of formulating various prompts and extracting knowledge captured by the PLMs extended with specific domain data. 

\section{GPT-2 Domain Adaptation}

The pretrained GPT-2 model was used for all of our experiments, which was then finetuned with News and Tweets corpus. Entity Characterization experiments is executed on both News and tweets domain adapted language models. Tweet Characterization experiments is executed on tweets domain adapted language model. The fine-tuning of GPT-2 PLM was carried out with default hyper-parameters set by Hugging-Face implementation.\footnote{HuggingfaceTransformers:\url{https://github.com/huggingface/transformers}}

\subsection{News Corpus}
GDELT is a global dataset that continuously monitors Broadcast, Print, and Web news worldwide in multiple languages and has recorded events from 2015 onward. The GDELT Database is integrated with Google Big-Query for fetching data. We downloaded news article URLs from the GDELT for seven different media houses and scraped the content of URLs. Each media house has its style of presenting content. Articles containing native language snippets were discarded and only the English language content was extracted, creating a corpus of articles for each Media House. Table \ref{tab:scrapedArticlesDetails} shows the count and size on disk of articles from each Media House. 

We finetune the pretrained GPT-2 (345M) on the news corpora from seven different media houses, training a separate model for each Media House.

\begin{table}[!htbp]
\centering
\caption{Scraped Articles from each Media House }
\label{tab:scrapedArticlesDetails}
\begin{tabular}{ccc} 
\toprule
\textbf{Media house} & \begin{tabular}[c]{@{}c@{}}\textbf{No. of}\\\textbf{ Articles}\end{tabular} & \begin{tabular}[c]{@{}c@{}}\textbf{Size of}\\\textbf{ Training Data (MB)}\end{tabular} \\ 
\toprule
Media House A & $41,968$                                                    & $283$            \\ 
\midrule
Media House B & $56,824$                                                   & $166$            \\ 
\midrule
Media House C & $18,012$                                                    & $75$             \\ 
\midrule
Media House D & $26,064$                                                    & $25$             \\ 
\midrule
Media House E & $34,284$                                                    & $85$             \\ 
\midrule
Media House F & $35,184$                                                    & $201$            \\ 
\midrule
Media House F & $95,676$                                                   & $208$            \\
\bottomrule
\end{tabular}
\end{table}

\begin{table}[!htbp]
    \centering
    \caption{Tweets Corpora}
    \label{tab:tweetFineTuningStats}
    \begin{tabular}{cc}
       \toprule
        \textbf{Dataset Category} & \textbf{Number of Tweets} \\
        \toprule
        Government Policy & 1,076,795 \\ 
        \midrule
        Economically Weaker Section Abuse & 88,416 \\
        \midrule
        Agriculturists Voice & 56,271 \\
        \bottomrule
    \end{tabular}
\end{table}

\subsection{Tweets Corpus}
We used a custom collated dataset from 3 different categories of social media movements for tweet characterization: \textit{Government Policy, Agriculturists' Voice, and Economically Weaker Section Abuse}. The data collected for our analysis spans over a period of $2$ years, and included prolonged heated discussions in the context of the related country. We used \textit{snowball sampling}~\cite{goodman1961snowball} of the trending hashtag to collect tweets of events for our experiment. The list of events and the collected data is listed in the Table~\ref{tab:tweetFineTuningStats}.
These datasets were collated during their peaks in their respective timelines from Twitter. All these movements resulted in quite a turmoil in the online verse and hence used for domain adaptation on informal texts. The data was extracted using standard Twitter developer APIs using the then trending hashtags. 

A lot of noise was present in the collated data from the online universe, it included multi-lingual and code-mixed data. Since, we wish to focus only on roman scripts and English tweets as part of our experiments, we cleaned the data.\footnote{Since GPT-2 PLM is trained on English corpus and our goal was to generate characterization in English language} Moreover, there was content in the dataset like URLs and user mentions which were not needed for characterization. In order to get the tweets ready for the model's domain adaptation, we passed the tweets through the following steps:
\begin{itemize}
    \item Lower case the tweets
    \item Remove user mentions
    \item Remove hashtags
    \item Replace the emojis with their \\ corresponding texts, surrounded by colons
    \item Remove punctuation
\end{itemize}

After completing the above steps, only Tweets with greater than 70\% of words found in the English dictionary are considered. Subsequently, we finetuned the pretrained GPT-2 (774M) model with the cleaned data. 

\section{Characterization with Designed Prompts}
The theory and approach we have pursued in this work are strongly aligned with the idea of \textit{``programming in natural language''} detailed by \citeauthor{reynolds2021prompt} \cite{reynolds2021prompt}. The language model would fail because ``the probability distribution produced in response to a prompt is not a distribution over ways \textit{a person would} continue that prompt, it’s the distribution over the ways \textit{any person could} continue that prompt''. Hence prompt design should constrain the entailment generation ( i.e. predicted sequence of the language model ) to produce the desired effect and circumvent the irrelevant. Task-agnostic prompts are less effective when compared to task-specific prompts. 

Two main types of prompts can be used as inputs for a language model, namely \textit{cloze prompts} and \textit{prefix prompts}. Cloze prompts as in \cite{petroni2019language} is where the ``answer'' token is masked. The model predicts the masked token, whereas in prefix prompts \cite{li2021prefix, lester2021power}, also known as priming, the language model acts as a sequence generator where a prompt is fed as input and left to the model to generate conditional text auto-regressively.

`The datasets GLUE \cite{wang2018glue} and SuperGLUE \cite{wang2019superglue} were, introduced by \citeauthor{wang2018glue} to evaluate \textit{natural language understanding} tasks. The datasets contain context, which has to be understood by the model and subsequently perform a task. We have a similar approach in two forms. 
One, we append designed conditional questions to tweets, suited for priming, to generate characterizing entailments. Two, a synopsis describes a particular characterization concept followed by the tweet and a conditional question, together forming a broad context to prime the language model in order to characterize the tweet. 

In this work we have designed \textit{conditional prefix prompts} for the \textit{``Entity Characterization''} task and \textit{detailed context and conditional question} for the \textit{``Tweet Characterization''} task. 

\subsection{Entity Characterization}
We experiment with text entailment, generated by domain-adapted GPT-2 PLMs for subjective characterization of the entity in the designed input. The models are finetuned with formal texts in news corpora and informal texts in tweets corpora. The input for an experiment consisted of a entity appended with a \textit{Designed Conditional Prefix-prompt}. Entailment text was generated using GPT-2 predicted tokens with the input and evaluated the effect of prefix-prompt on the entailment text to characterize entity subjectively. We considered raw outputs for evaluation, and we did not post-process to create a proper sentence. Instead, we evaluate raw output to check if it contains adjectives relevant and well-known for the corresponding entity.

News articles are written with a reasonable research and insights hence, we consider them as formal text. Tweets are general thoughts shared with limited or good insights from a large group of users hence, are considered as informal texts.

First, for entity characterization with formal text corpus, we experiment with entities in news corpus articles. The experiments considered the top 10 entities from all the seven media houses and eight designed prefix prompts shown in Table \ref{tab:designedPrefixPrompts}. 
Ten outputs were generated for each media house per (entity, prefix-prompt) pair, and evaluated.

\begin{table}[!htbp]
\caption{Designed Conditional Prefix-prompts}
\label{tab:designedPrefixPrompts}
\begin{tabular}{c|c} 
\toprule
``is a very''             & ``is known as''           \\ 
\midrule
``can be described as a'' & ``is regarded as a''      \\ 
\midrule
``lacks''                 & ``is called the''         \\ 
\midrule
``probably is a''         & ``can be inferred as a''  \\
\bottomrule
\end{tabular}
\end{table}

The GPT-2 PLM is finetuned with vocabulary $\mbox{\large $V$}$ from a single media house news corpus or formal domain $\mbox{\large $FD$}$ to create a domain adapted language model $\mbox{\large $M_{FD}$}$. With $\mbox{\large $M_{FD}$}$, raw entailments are generated for a pair of entity $\mbox{\large $E$}$ and designed conditional prefix as shown in Table \ref{tab:designedPrefixPrompts}. Following are examples of input pairs:
$$\mathrm{x1} = (\text { E, \textbf{\textit{is a very}}})$$
$$\mathrm{x2} = (\text { E, \textbf{\textit{can be described as a}}})$$

Considering entailments as characterizing entities, entity characterization $\mbox{\large $CH$}$ can be shown as:
$$
CH(\mathrm{x1, M_{FD}}) = \text {Jane \textbf{\textit{is a very}} } \rule{1cm}{0.15mm}
$$
$$
CH(\mathrm{x2, M_{FD}}) = \text {Jane \textbf{\textit{can be described as a}} } \rule{1cm}{0.15mm}
$$

One domain adapted language model $\mbox{\large $M_{FD}$}$ is created for each of the seven media houses corpora as shown in the Table \ref{tab:scrapedArticlesDetails} and subsequently, we conduct experiments on each of them. 
We evaluate the performance of prefix prompts in input $\mbox{\large $x$}$ to characterize entity $\mbox{\large $E$}$ by generated entailment in $\mbox{\large $CH(\mathrm{x, M_{FD}})$}$ with formal text-domain $FD$ adapted language model $\mbox{\large $M_{FD}$}$.

Second, entity characterization with tweets corpus or informal domain adapted language model $\mbox{\large $M_{ID}$}$ is created with a category of tweets as shown in Table \ref{tab:tweetFineTuningStats} considered to be informal text domain $\mbox{\large $ID$}$. Considering language model adapted with informal texts generate characterization of entities, characterization of entities $\mbox{\large $CH$}$ can be shown as:
$$
CH(\mathrm{x1, M_{ID}}) = \text {Jane \textbf{\textit{is a very}} } \rule{1cm}{0.15mm}
$$
$$
CH(\mathrm{x2, M_{ID}}) = \text {Jane \textbf{\textit{can be described as a}} } \rule{1cm}{0.15mm}
$$
One domain adapted language model $\mbox{\large $M_{ID}$}$ is created for each of the three tweet categories corpora as shown in Table \ref{tab:tweetFineTuningStats}.
Four most frequently appearing entities across the corpora were chosen for the experiments. Ten outputs were generated for each (entity, prefix-prompt) pair on each model, and an evaluation was done on results with higher number of adjectives. We evaluated the performance of prefix prompts in input $\mbox{\large $x$}$ to generate characterization of entity $\mbox{\large $E$}$ by generated entailment in $\mbox{\large $CH(\mathrm{x, M_{ID}})$}$ with informal text-domain $ID$ adapted language model $\mbox{\large $M_{ID}$}$.

\subsection{Tweet Characterization}
To understand the effectiveness of contextual prompts on tweet characterization, we focus on the domain adaptation of a pretrained language model. The model is finetuned on social media corpora as shown in Table \ref{tab:tweetFineTuningStats}, and prompted to generate outputs to mine commonsense knowledge and reasoning stored in the language model.

For the first set of experiments, as shown in Table \ref{tab:my-table1}, we exploit BoolQ \cite{clark2019boolq} dataset format, where we first give a tweet as context followed by a Yes/No format question. The posed question try to exploit the commonsense knowledge stored in the GPT-2 model as shown by the authors~\cite{GPT2}. 
Our main aim is to investigate high-level concepts like advocacy, hyper-advocacy, disinformation, propaganda and stance. These were all explicitly asked in the corresponding questions.

\begin{table*}[!htbp]
\centering
\caption{Tweet Characterization by \textbf{``Commonsense Reasoning''} similar to BoolQ \cite{clark2019boolq}}
\begin{tabular}{l} 
\toprule
\multicolumn{1}{c}{\textbf{Type 1 Experiment Designed Templates}}                                                                       \\ 
\toprule
\textless{}\textit{Tweet}\textgreater{}. Q: Is it true that preceding sentence advocates a cause? A: \textless{}\textit{True or False or Subjective Text}\textgreater{}          \\ 
\midrule
\textless{}\textit{Tweet}\textgreater{}. Q: Is it true that preceding sentence hyper-advocates a cause ? A: \textless{}\textit{True or False or Subjective Text}\textgreater{}   \\ 
\midrule
\textless{}\textit{Tweet}\textgreater{}. Q: Is it true that preceding sentence is a disinformation ? A: \textless{}\textit{True or False or Subjective Text}\textgreater{}       \\ 
\midrule
\textless{}\textit{Tweet}\textgreater{}. Q: Is it true that preceding sentence is a about a propaganda ? A: \textless{}\textit{True or False or Subjective Text}\textgreater{}   \\ 
\midrule
\textless{}\textit{Tweet}\textgreater{}. Q: Is it true that preceding sentence favors a cause ? A: \textless{}\textit{True or False or Subjective Text}\textgreater{}            \\ 
\midrule
\textless{}\textit{Tweet}\textgreater{}. Q: Is it true that preceding sentence is against a cause? A: \textless{}\textit{True or False or Subjective Text}\textgreater{}         \\ 
\midrule
\textless{}\textit{Tweet}\textgreater{}. Q: The preceding statement is advocating a cause. True or False? A: \textless{}\textit{True or False or Subjective Text}\textgreater{}  \\
\bottomrule
\end{tabular}

\label{tab:my-table1}
\end{table*}

Next, we employ multiple choice question (MCQ) format, similar to AG's News\footnote{AG News Dataset: \url{http://goo.gl/JyCnZq}}, expecting a one-word or an option as a predicted output from the domain adapted PLM (Table \ref{tab:my-table2}). We aim to extract commonsense knowledge from the generative model given a task description \cite{GPT2}.

\begin{table}[!htbp]
\centering
\caption{Tweet Characterization by \textbf{``MCQ''} similar to AG's News}
\begin{tabular}{l} 
\toprule
\multicolumn{1}{c}{\textbf{Type 2 Experiment Designed Templates}}                                                                                                                                                                                                                                                 \\ 
\toprule
\begin{tabular}[c]{@{}l@{}}\textless{}\textit{Tweet}\textgreater{}. Q: Can the preceding sentence be classified as\\one of the following: information, rhetoric, advocacy,\\hyper-advocacy, dogma, or propaganda?\\A: \textless{}\textit{One of the listed options or Subjective} Text\textgreater{}\end{tabular}  \\ 
\midrule
\textless{}\textit{Tweet}\textgreater{}. Classify the preceding sentence as one of the following:                                                                                                                                                                                                                                  \\
A) information                                                                                                                                                                                                                                                                                  \\
B) disinformation                                                                                                                                                                                                                                                                               \\
C) advocacy                                                                                                                                                                                                                                                                                     \\
D) hyper-advocacy                                                                                                                                                                                                                                                                               \\
E) propaganda                                                                                                                                                                                                                                                                                   \\
F) none of the above                                                                                                                                                                                                                                                                            \\
\textless{}\textit{One of the above option or Subjective Text}\textgreater{}                                                                                                                                                                                                                                                      \\
\bottomrule
\end{tabular}
   
\label{tab:my-table2}
\end{table}

Subsequently, our next set of experiments, as shown in Table \ref{tab:my-table3}, focuses on asking the PLM to classify a given tweet into specific low-level concepts. The low-level concepts, we test our model on, include the classes/concepts which are not domain-specific like advocacy and hyper-advocacy, but rather standard terms in articles and communicative english. For example, call-to-action, dogma, EML, emotion, sentiment and propaganda. This experiment also tests the model for text classification, knowledge mining, and sentiment analysis.

\begin{table}[!htbp]
\centering
\caption{Tweet Characterization by \textbf{``General Commonsense Reasoning''} similar to MultiRC \cite{khashabi2018looking}}.
\begin{tabular}{l} 
\toprule
\multicolumn{1}{c}{\textbf{Type 3 Experiment Designed Template}}                                       \\ 
\toprule
\textless{}\textit{Tweet}\textgreater{}. Q:~For~what~cause~is~the~above~Tweet~CTA?\\A:~\textless{}\textit{Subjective Text}\textgreater{}                   \\ 
\midrule
\textless{}\textit{Tweet}\textgreater{}. Q:~For~what~cause~is~the~above~Tweet~call-to-action?\\A:~\textless{}\textit{Subjective Text}\textgreater{}        \\ 
\midrule
\textless{}\textit{Tweet}\textgreater{}. Q:~Do~you~find~the~above~Tweet~as~having\\dogmatic~content?\\A: \textless{}\textit{Subjective Text}\textgreater{}   \\ 
\midrule
\textless{}\textit{Tweet}\textgreater{}. Q:~Do~you~find~the~above~Tweet~as~having\\rhetoric~content?\\A:~\textless{}\textit{Subjective Text}\textgreater{}  \\ 
\midrule
\textless{}\textit{Tweet}\textgreater{}. Q:~Is~the~Tweet~call~to~action~for~a~cause/protest?\\A:~\textless{}\textit{Subjective Text}\textgreater{}         \\ 
\midrule
\textless{}\textit{Tweet}\textgreater{}. Q:~What~is~the~dominant~emotion~in~the~above~\textless{}\textit{Tweet}\textgreater{}?\\A:~\textless{}\textit{Subjective Text}\textgreater{}         \\ 
\midrule
\textless{}\textit{Tweet}\textgreater{}. Q:~What~is~the~sentiment~of~the~above~\textless{}\textit{Tweet}\textgreater{}?\\A:~\textless{}\textit{Subjective Text}\textgreater{}                \\ 
\midrule
\textless{}\textit{Tweet}\textgreater{}. Q:~Is~the~above~Tweet~a~propaganda?\\A:~\textless{}\textit{Subjective Text}\textgreater{}                         \\
\bottomrule
\end{tabular}

\label{tab:my-table3}
\end{table}

Our last set of experiments include prompts similar to Reading Comprehension (RC) where the questions are based on a given passage, as shown in Table \ref{tab:my-table4}. For example, each prompt starts with defining a concept like dogmatism, followed by a few examples that can be classified as having dogmatic content and posing a question for the model to generate answers to, building on the grounds as shown by \citeauthor{zhang2018record}\cite{zhang2018record}. We further tone down our question domain by mentioning the set of emotions to classify into dogmatic content, call-to-action and asking a yes/no question for emotionally manipulative language.

Our primary aim in all the above experiments was to be able to test and analyze the GPT-2 model for tasks which are an intersection of Commonsense Knowledge Mining (CKM), Commonsense Reasoning (CR), Text classification (TC) and Question Answering (QA). Given a specific task description without any transfer learning, the model should generate expected/logical outputs as claimed by the authors in \cite{GPT2}.

\begin{table*}[!htbp]
\centering
\caption{Tweet Characterization by \textit{``Larger-context Commonsense Reasoning''} similar to ReCoRD \cite{zhang2018record}}
\begin{tabular}{l} 
\toprule
\multicolumn{1}{c}{\textbf{Type 4 Experiment Designed Template}}                                                                                                                           \\ 
\toprule
\textit{\textless{}Descriptive Synopsis on Dogmatism\textgreater{}}                                                                                                       \\
Question: John says "\textit{\textless{}Tweet\textgreater{}}". Does John’s saying contain dogmatic content?                                                               \\ 
\midrule
\textit{\textless{}Descriptive Synopsis on Emotionally Manipulative Language\textgreater{}}                                                                               \\
Question: John says "\textit{\textless{}Tweet\textgreater{}}". Does John’s saying contain Emotionally Manipulative Language?                                              \\ 
\midrule
\textit{\textless{}Descriptive Synopsis on Emotional Analysis. Definitions of Aggressive, Optimism, Love, Submission, Fear, Surprise, Sadness and Disgust\textgreater{}}  \\
Question: Jhon says "\textit{\textless{}Tweet\textgreater{}}". Does Jhon’s saying contain Aggressive/Optimism/Love/Submission/Fear/Surprise/Sadness/Disgust emotion?      \\ 
\midrule
\textit{\textless{}15 Definitions of Call-to-action followed by example texts\textgreater{}}                                                                              \\
Question: John says "\textit{\textless{}Tweet\textgreater{}}". Can John’s saying be classified as Call-To-Action?                                                         \\
\bottomrule
\end{tabular}

\label{tab:my-table4}
\end{table*}

\section{Results}
\subsection{Entity Characterization with News Datasets}
Generating entailments is similar to \textit{Natural Language Generation}. With designed prefix-prompts we are interested in evaluating the quality of free-form text generation which characterizes an entity. Hence, we evaluated the efficiency of prefix-prompts to generate relevant and entity characterizing entailments with following measures:
\begin{enumerate}
  \item Iterations needed to generate ten valid English sentence entailments - Table \ref{tab:syntaxFail}
  \item The ratio of entailments of negative and positive sentiment for each prefix-prompt across media houses - Table \ref{tab:sentimenttotal}
  \item Percentage of the positive sentiment of each entity  in each media house - Table \ref{tab:sentimentMediaHouse}
  \item Presence of adjective POS tags in entailments. Table \ref{tab:adjPresent}.
  \item Human evaluation of entity relevant and characteristic entailments - Table \ref{tab:relanddesc} and \ref{tab:relEntity}
  \item Cluster analysis of outputs with Sentence Embeddings
\end{enumerate}

GPT-2 fine-tuning datasets had noise content like hashtags, emoji, and others. Iterations were done to get clear English sentences. Table \ref{tab:syntaxFail} shows the number of failed outputs it took to generate ten good outputs across all seven Media houses. The \textit{``is a very''} and \textit{``is regarded as''} prefix-prompts required lesser iterations compared to \textit{``lacks''} and \textit{``is called the''} prefix-prompts.

\begin{table}[!htbp]
\centering
\caption{Failed outputs per Prefix-prompt}
\label{tab:syntaxFail}
\begin{tabular}{cc}
\toprule
\textbf{Prefix Prompts} & \textbf{Fail Count}\\
\toprule
\underline{\textbf{is a very}} & \underline{\textbf{89}}\\
\midrule
is known as & 189\\
\midrule
can be described as a & 93\\
\midrule
\underline{\textbf{is regarded as a}} & \underline{\textbf{60}}\\
\midrule
\textit{\textbf{lacks}} & \textit{\textbf{287}}\\
\midrule
\textit{\textbf{is called the}} & \textit{\textbf{349}}\\
\midrule
probably is a & 141\\
\midrule
can be inferred as a & 126\\
\bottomrule
\end{tabular}
\end{table}

The Sentiment of all entailments was analyzed for sentiment analysis using AllenNLP Sentiment Analyzer\footnote{Allen NLP Sentiment Analysis: \url{https://demo.allennlp.org/sentiment-analysis/glove-sentiment-analysis}}. As shown in Table \ref{tab:sentimenttotal}, the \textit{``is regarded as a''}, \textit{``is Known as''} and \textit{``is called the''} are two most positive sentiment generating prefix-prompts while \textit{``lacks''} is the most negative sentiment generating prefix-prompt.

\begin{table}[!htbp]
\centering
\caption{Sentiment of entailments for each Prefix-prompt}
\label{tab:sentimenttotal}
\setlength{\tabcolsep}{2pt} 
\renewcommand{\arraystretch}{0.8} 
\begin{tabular}{cccc}
\toprule
\begin{tabular}[c]{@{}c@{}}\textbf{Prefix}\\\textbf{ Prompts}\end{tabular} & \begin{tabular}[c]{@{}c@{}}\textbf{Negative}\\\textbf{ Sentiments}\end{tabular} & \begin{tabular}[c]{@{}c@{}}\textbf{Positive}\\\textbf{ Sentiments}\end{tabular} & \begin{tabular}[c]{@{}c@{}}\textbf{\%age of}\\\textbf{Positive}\\\textbf{Outputs} \end{tabular}\\
\toprule
\underline{\textbf{is a very}} & 34 & 666 & \underline{\textbf{95.14}} \\
\midrule
\underline{\textbf{is known as}} & 21 & 679 & \underline{\textbf{97.00}}\\
\midrule
can be described as a & 37 & 663 & 94.71\\
\midrule
\underline{\textbf{is regarded as a}} & 17 & 683 & \underline{\textbf{97.57}}\\
\midrule
\textbf{\textit{lacks}} & 375 & 325 & \textit{\textbf{46.43}}\\
\midrule
\underline{\textbf{is called the}} & 21 & 679 & \underline{\textbf{97.00}}\\
\midrule
probably is a & 110 & 590 & 84.29 \\
\midrule
can be inferred as a & 30 & 670 & 95.71\\
\bottomrule
\end{tabular}
\end{table}

Table \ref{tab:sentimentMediaHouse} shows the percentage of positive sentiment entailments for each entity for each Media house. For some entities, the number of total positive sentiments was consistently low, whereas, for some entities, it was high. Media houses \textit{M1 M2 and M6} produced more positive sentiments towards \textit{P1 P6 and P10} entities, who belong to the same Political Party.

\begin{table}
\centering
\caption{Positive Sentiment percentage of Entities across Media Houses}
\label{tab:sentimentMediaHouse}
\begin{tabular}{cccccccc} 
\toprule
\multirow{2}{*}{\begin{tabular}[c]{@{}c@{}}\textbf{Entity}\end{tabular}} & \multicolumn{7}{c}{\textbf{Media Houses}}                                                                      \\ 
\cmidrule{2-8}
                                                                                          & M2            & M7            & M5            & M1            & M4            & M3            & M6             \\ 
\toprule
P1                                                                                        & \textbf{92.5} & 91.3          & 88.8          & 92.5          & 90            & 91.3          & \textbf{95}    \\
P2                                                                                        & \textbf{93.8} & 90            & 90            & 88.8          & \textbf{92.5} & 87.5          & 85             \\
P3                                                                                        & 87.5          & \textbf{92.5} & \textbf{93.8} & \textbf{93.8} & 87.5          & 85            & 86.3           \\
P4                                                                                        & 87.5          & \textbf{83.8} & 87.5          & 86.3          & \textbf{92.5} & 87.5          & 87.5           \\
P5                                                                                        & 87.5          & 91.3          & \textbf{92.5} & 90            & 88.8          & \textbf{93.8} & \textbf{93.8}  \\
P6                                                                                        & 91.3          & \textbf{93.8} & 90            & 88.8          & 90            & 91.3          & \textbf{93.8}  \\
P7                                                                                        & \textbf{85}   & \textbf{76.3} & \textbf{82.5} & \textbf{82.5} & \textbf{78.8} & 85            & 85             \\
P8                                                                                        & \textbf{83.8} & \textbf{82.5} & \textbf{81.3} & \textbf{82.5} & \textbf{78.8} & \textbf{80}   & \textbf{82.5}  \\
P9                                                                                        & 88.8          & 90            & 87.5          & 83.8          & \textbf{93.8} & 88.8          & \textbf{92.5}  \\
P10                                                                                       & \textbf{96.3} & 86.3          & 90            & \textbf{93.8} & 91.3          & 91.3          & 90             \\
\bottomrule
\end{tabular}
\end{table}

Usually, to praise or criticize someone, adjectives are used. All entailments for each prefix-prompt are analyzed with the help of NLTK POS tagging\footnote{NLTK TAG: \url{https://www.nltk.org/api/nltk.tag.html}} to check the presence of Adjectives POS Tags(JJ), Superlative Adjective POS Tags(JJS), and Comparative Adjective POS Tags(JJR). Table \ref{tab:adjPresent} shows the presence of above mentioned POS tags in 700 entailments for each prefix-prompt. "is a very" and "can be described as" prefix-prompts have most cases adjective POS tags while "is known as" and "is called the" prefix-prompts have least outcomes with Adjective POS tags.

\begin{table}[]
\centering
\caption{Adjective POS tags in 700 Characterization Outputs}
\label{tab:adjPresent}
\begin{tabular}{ccc}
\toprule
    \textbf{Prefix-prompt}&
  \textbf{POS Absent} &
  \textbf{POS Present} \\ 
  \toprule
\underline{\textbf{is   a very}}             & 152  & \underline{\textbf{548}}\\
\midrule
\textit{\textbf{is   known as}}           & 328  & \textit{\textbf{372}}\\
\midrule
\underline{\textbf{can be described as a}}   & 162  & \underline{\textbf{538}}  \\
\midrule
is   regarded as a      & 169  & 531  \\
\midrule
lacks                   & 276  & 424  \\
\midrule
\textit{\textbf{is called the}}         & 351  & \textit{\textbf{349}}  \\
\midrule
probably   is a         & 276  & 424    \\
\midrule
can   be inferred as a  & 181  & 519 \\ 
\bottomrule
\end{tabular}
\end{table}

\textbf{Human evaluation} was done on the following two attributes with one human evaluator for each Media House.   
\begin{enumerate}
    \item If the entailment is relevant to the entity and factually correct.
    \item If the entailment is valid and describes the characteristic of the entity.
\end{enumerate}
Another attribute to mark for the output was the correctness of the characteristics described by the entailment. Entity characterizations are cognitive qualities of a person and are subject more to individual perception. Therefore, the evaluator's perception of the entity could greatly bias his opinion to validate the correctness of characterization in the entailment. Because of the mentioned reason, the third attribute should only be marked by a Domain expert.

Table \ref{tab:relanddesc} shows that out of total outputs, 35.98\% were relevant to the entities. This score is not close to the 50\% benchmark for GPT-2. However, since the score of 35.98\% is only on a short experiment, it tends to improve over a large number of experiments. Out of total relevant output, 74\% of outputs describe characteristics of entities. So we can say the efficiency of prefix-prompts is 74\% to get characteristic details about an entity.

By comparing the individual prefix-prompts for the number of relevant as well as characterizing outputs, as shown in the Table \ref{tab:relEntity}, \textit{``is a very''} is the best performing prefix-prompt. It has 55.4\% entity relevant entailments and 81.71\% entity characterizing entailments. On the other hand, \textit{``is called the''} prefix-prompt is the least performing on both the scales.

\begin{table}[!htbp]
\centering
\caption{Relevant and characterizing outputs generated }
\label{tab:relanddesc}
\setlength{\tabcolsep}{1.3pt} 
\renewcommand{\arraystretch}{0.8} 
\begin{tabular}{ccccc}
\toprule
& \begin{tabular}[c]{@{}c@{}}\textbf{Non- Relevant}\\\textbf{Output}\end{tabular}& \begin{tabular}[c]{@{}c@{}}\textbf{Only}\\\textbf{Relevant} \end{tabular} & \begin{tabular}[c]{@{}c@{}}\textbf{Relevant \&}\\\textbf{Characterizing}\end{tabular} & \begin{tabular}[c]{@{}c@{}}\textbf{Total}\\\textbf{Relevant} \end{tabular} \\
\toprule
\begin{tabular}[c]{@{}c@{}}\textbf{No. of}\\\textbf{Outputs}\end{tabular}           & 3585             & 528                        & 1487                     & 2015           \\
\midrule
\begin{tabular}[c]{@{}c@{}}\textbf{\%age of}\\\textbf{Outputs}\end{tabular} & 64.02            & 9.43                       & 26.55                    & \underline{\textbf{35.98}} \\
\bottomrule
\end{tabular}
\end{table}

\begin{table}[!htbp]
\centering
\caption{Percentage of Relevant and Characterizing outputs for each Prefix-prompt for all Media Houses }
\label{tab:relEntity}
\setlength{\tabcolsep}{1.3pt} 
\renewcommand{\arraystretch}{0.8} 
\begin{tabular}{cccc}
\toprule
\begin{tabular}[c]{@{}c@{}}\textbf{Prefix}\\\textbf{Prompts}\end{tabular} & \begin{tabular}[c]{@{}c@{}}\textbf{Relevant \&}\\\textbf{Characterizing}\\\textbf{Entities}\end{tabular} & \begin{tabular}[c]{@{}c@{}}\textbf{Relevant} \\\textbf{to}\\\textbf{Entities}\end{tabular} & \begin{tabular}[c]{@{}c@{}}\textbf{\%age}\\\textbf{ Characterizing}\\\textbf{Entities}\end{tabular} \\
\toprule
\underline{\textbf{is a very}}                  & 45.29 & \underline{\textbf{55.43}} & \underline{\textbf{81.71}} \\
\midrule
is known as             & 25    & 33.43 & 74.78 \\
\midrule
can be   described as a  & 29.43 & 38.57 & 76.30 \\
\midrule
is regarded as a         & 31    & 43.14 & 71.86 \\
\midrule
lacks                    & 25.71 & 33.43 & 76.91 \\
\midrule
\underline{\textit{\textbf{is called the}}}            & 15.71 & \underline{\textit{\textbf{24.57}}} & \underline{\textit{\textbf{63.94}}} \\
\midrule
probably is a            & 19.86 & 29.43 & 67.48 \\
\midrule
can be inferred as a      & 20.43 & 29.86 & 68.42\\
\bottomrule
\end{tabular}
\end{table}

\textbf{Cluster analysis of outputs}. Sentence Transformers can be used to calculate sentence embeddings for each output statement. We use pretrained Sentence-BERT (SBERT) embeddings \footnote{ \url{https://www.sbert.net/}} to calculate 768 length vector for each output, irrespective of the sentence length. Sentence Embedding vectors represent complete sentence with more focus on the context. This ability of Sentence Embeddings makes it more useful to evaluate entity characterization entailments.

k-means clustering algorithm is used to group similar sentence vectors together. Clusters are then validated against the human annotations and other results as shown in Table \ref{tab:clusterAnalysis}. Optimal clusters were computed based on Distortion, Silhouette and Calinski-harabaz scores. Distortion is  ``the sum of the squared distances between each observation vector in cluster and its dominating centroid''. Silhouette score ``quantifies to what extent a given cluster is cohesive and separate''. Calinski-Harabasz score ``is the ratio of the sum of between-clusters dispersion and of inter-cluster dispersion for all clusters''. On comparing the scores across the metrics, Silhouette score performed the best. Hence, $k=4$ is chosen based on the Silhouette score. 

Table \ref{tab:clusterAnalysis} shows the count of output in clusters across Sentiment, Adjective and Relevance dimensions. Following are key observations from Cluster Analysis:
\begin{enumerate}
    \item Cluster 0 has Relevant and Characterizing Entity outputs
    \item Cluster 1 has maximum negative sentiment outputs
    \item Cluster 3 has maximum adjective absent outputs
    \item Cluster 3 has maximum irrelevant outputs
    \item All clusters have high positive sentiment outputs
\end{enumerate}
The key observation is in Cluster 3, where there are maximum adjectives absent and irrelevant outputs.

\begin{table*}[!h]
\centering
\caption{Cluster Analysis of Outputs}
\label{tab:clusterAnalysis}
\setlength{\tabcolsep}{7pt} 
\renewcommand{\arraystretch}{1} 
\begin{tabular}{cccccccc}
\toprule
\textbf{Cluster} & \begin{tabular}[c]{@{}c@{}}\textbf{Negative}\\\textbf{Sentiment}\end{tabular} & \begin{tabular}[c]{@{}c@{}}\textbf{Positive}\\\textbf{Sentiment}\end{tabular} & \begin{tabular}[c]{@{}c@{}}\textbf{Adjective}\\\textbf{Absent}\end{tabular} & \begin{tabular}[c]{@{}c@{}}\textbf{Adjective}\\\textbf{Present}\end{tabular} & \textbf{Irrelevant} & \begin{tabular}[c]{@{}c@{}}\textbf{Only Relevant}\end{tabular} & \begin{tabular}[c]{@{}c@{}}\textbf{Relevant and} \\\textbf{Characterize Entity}\end{tabular}
\\
\toprule
0       & 98            & 1381      & 323        & \textbf{1156}             & 690  & 143 & \textbf{646}          \\ \midrule
1  & \textbf{283}           & 749      & 305        & 727           & 683  & 82  & 267           \\ \midrule
2       & 145           & 1279       & 315        & \textbf{1109}             & 966  & 173 & 285          \\ \midrule
3       & 119           & 1546       & \underline{\textbf{952}}        & 713            & \underline{\textbf{1246}} & 130 & 289         \\
\bottomrule
\end{tabular}
\end{table*}
To add to the above results, we repeated all the experiments on off-the-shelf GPT-2 PLM and noticed that generated entailments were incoherent, too random, and insensible. 

\textbf{The critical takeaway} from entity characterization evaluation is that \textbf{``is a very''} prefix-prompt is performing the best across all output analysis and characterizing entities with formal-text domain adapted GPT-2 PLM.

\begin{table}[h]
\centering
\caption{Annotation results upon agreement between the domain experts }
\label{tab:tweetEntityAgreement}
\setlength{\tabcolsep}{4pt} 
\renewcommand{\arraystretch}{0.8} 
\begin{tabular}{ccc}
\toprule
  \textbf{Model Used}  &
  \begin{tabular}[c]{@{}c@{}}\textbf{Agreement}\\\textbf{(Cohen's kappa)}
  \end{tabular}  
  & \begin{tabular}[c]{@{}c@{}}\textbf{\%age} \\\textbf{Characterizing}\\\textbf{Outputs} 
  \end{tabular}  
  \\ 
  \toprule
   Government Policy            & 0.94   & 47.65 \\
\midrule
   \begin{tabular}[c]{@{}c@{}}Economically Weaker\\Section Abuse 
  \end{tabular} & 0.74 & 72.65\\
\midrule
   Agriculturists Voice & 0.66  & 60.15\\
\midrule
  Vanilla GPT-2      & 0.87 & 49.21\\
\bottomrule
\end{tabular}
\end{table}

\begin{table*}[!h]
\centering
\caption{Effects of Prefix-prompts on Tweet Corpus Entities}
\label{tab:prefixPromptPerformance}
\setlength{\tabcolsep}{2pt} 
\renewcommand{\arraystretch}{0.8} 
\begin{tabular}{ccccccccc} 
\toprule
\multirow{2}{*}{\begin{tabular}[c]{@{}c@{}}\textbf{Analysis}\\\textbf{Measures}\end{tabular}} & \multicolumn{7}{c}{\textbf{Prefix-prompts}}                                                                      \\ 
\cmidrule{2-9}

 & ``can be described as a'' & ``can be inferred as'' & ``is a very'' & ``is called the'' & ``is known as'' & ``is regarded as a'' & ``lacks''   & ``probably is a''  \\
\toprule                                 
\begin{tabular}[c]{@{}c@{}}\textbf{adjectives}\\\textbf{centroid}\\\textbf{ distance}\end{tabular}  & 0.64                  & 0.73               & 0.57      & 0.77          & 0.67        & 0.72             & 0.62    & 0.68           \\
\midrule
\begin{tabular}[c]{@{}c@{}}\textbf{SBERT}\\\textbf{centroid}\\\textbf{ distance}\end{tabular} & 0.51                  & 0.50               & 0.43      & 0.52          & 0.50        & 0.51             & 0.51    & 0.44           \\
\midrule
\textbf{\%age adjectives} & 79.17               & 79.17            & 93.75   & 83.33       & 97.92     & 89.58          & 95.83 & 77.08        \\
\midrule
\begin{tabular}[c]{@{}c@{}}\textbf{\%age positive}\\\textbf{sentiment}\end{tabular}
 & 95.83               & 85.42            & 87.50   & 93.75       & 91.67     & 97.92          & 60.42 & 91.67        \\
\midrule
\begin{tabular}[c]{@{}c@{}}\textbf{\%age negative}\\\textbf{sentiment}\end{tabular}
            & 4.17                & 14.58            & 12.50   & 6.25        & 8.33      & 2.08           & 39.58 & 8.33         \\
\midrule
\begin{tabular}[c]{@{}c@{}}\textbf{\%age} \\\textbf{characterizing}\\\textbf{output}\end{tabular}
       & 64.58                & 50            & 60.4   & 56.25        & 62.5      & 70.83           & 54.16 & 58.3         \\
\bottomrule
\end{tabular}
\end{table*}

\subsection{Entity Characterization with Tweets Datasets}
We take the top four outputs for each (entity, prefix-prompt) pair, based on the adjectives present. They were then evaluated by two domain experts.
The criteria of the annotation was to validate whether the generated entailments talk about the information relevant to the entity, while characterizing the same.
Table \ref{tab:tweetEntityAgreement} shows the Cohen's kappa score for the inter-annotation agreement for each of the dataset adapted GPT-2 outputs. We also generated outputs with Vanilla GPT-2 model to quantify the effect of adaptation. The scores across all the four model outputs signify more than substantial agreement ( Cohen's kappa score > 0.6 ) between the two experts. We found that the generated characterizations were not only relevant but were also adapted to the specific movement on which the model had been finetuned. This included the majority opinion about the entity with respect to the particular dataset. Whereas, the vanilla GPT-2 PLM generated outputs relevant to the entity in general sense.

Since the characterization is often done using adjectives, we extracted the adjectives in each output and aggregated to show the significance of the characterization in outputs. To infer the effect of adaptation, we computed the Word2Vec centroid distance between the adjective sets of the adapted models and vanilla model outputs. On the similar lines, we also computed the SBERT centroid distances of outputs across the adapted models and the vanilla model outputs. Next, we calculated the percentage of outputs containing adjectives and the sentiment of outputs.
The centroid distances are greater than zero signifying outputs hence produced are learnt from the training datasets with high percentage of outputs containing adjectives.  To verify the effect of prefix-prompts, we generate outputs only with entities without the custom tailored prompts, and the observation by the experts is that the entailments were random and were not characterizing the entity. Table \ref{tab:prefixPromptPerformance} summarizes the analysis for each prefix-prompt.

Based on the user-mentions present in the tweets, we pick two most frequent entities each from the ruling party and the opposition party. Entities 1 and 2 are from the ruling party, and Entities 3 and 4 are from the opposition party. 
The model trained on the "Government Policy" dataset produced most varied sentiment outputs across all the entities, as shown in Table \ref{tab:tweetEntitySentiment}. 

\textbf{The key takeaway} is that for the informal text corpora, the finetuned models characterize the entities best with the prefix-prompt \textbf{``is regarded as a''} and worst with \textbf{``can be inferred as''}.

\begin{table}[H]
\centering
\caption{Positive Sentiment Outputs for each Entity across corpora}
\label{tab:tweetEntitySentiment}
\setlength{\tabcolsep}{4pt} 
\renewcommand{\arraystretch}{0.8} 

\begin{tabular}{cccc} 
\toprule
\multirow{3}{*}{\begin{tabular}[c]{@{}c@{}}\textbf{Entity}\end{tabular}} & \multicolumn{3}{c}{\textbf{Tweets Corpora}}                                                                      \\ 
\cmidrule{2-4}

 & \begin{tabular}[c]{@{}l@{}}Agriculturists \\Voice\end{tabular} & \begin{tabular}[c]{@{}l@{}}Government \\Policy\end{tabular} & \begin{tabular}[c]{@{}l@{}}Economically Weaker\\Section Abuse\end{tabular}  \\
 \toprule
Entity 1 & 90.63\%                                                        & 87.50\%                                                     & 96.88\%                                                                     \\ \midrule
Entity 2 & 96.88\%                                                        & 84.38\%                                                     & 90.63\%                                                                     \\ \midrule
Entity 3 & 90.63\%                                                        & 71.88\%                                                     & 93.75\%                                                                     \\ \midrule
Entity 4 & 84.38\%                                                        & 78.13\%                                                     & 90.63\%                                                                     \\
\bottomrule
\end{tabular}
\end{table}

\subsection{Tweet Characterization Results}
The four sets of experiments focused on analysing generative PLMs as text classifiers with abstract labels like dogma, advocacy, and hyper-advocacy. Starting with the first set of experiments, each subsequent one resulted from the failure of the previous set. We started with a basic true/false based question-answer format where every related question specifically stated the label, classifying the given sentence/tweet. 

On not getting the expected results from the generative language model, we tweaked the format to MCQ style question-answer format. All the labels into which the classification was needed were stated as different options. The experiments failed to expect a one-letter answer or even a one-word answer as the outputs were no way near even an "inferred" answer. 

Subsequently, we devised a new set of questions where we resorted to a bit low-level classification. The model is now asked to classify into general concepts usually used and shared in English. However, again, the answers were very subjective and not near the answer that can be expected from the posed question(s). 

Lastly, the fourth set of experiments includes introducing a low-level general concept and its definition, and a few related examples. Hence, forming a reading comprehension format followed by a question asking whether a given tweet can be classified as the introduced concept. Nevertheless, the model produced outputs that did not make sense with the question asked and was not valuable for the task defined in the prompt.

From all the failures to characterize the tweets, we concluded that the model, even though trained on 8 million webpages of formal text plus the huge corpus of informal text, failed to classify the outputs into abstract terms like dogma. This shows, that more formal supervised training needs to be performed for the model to comprehend those abstract labels.

\section{Conclusions and Future Work}
With zero-shot, we have observed that well-designed linguistic prompts with finetuned language models lead to good results on entity characterization. With this work, we attempt to solve NLP tasks in a zero-shot setting by exploring linguistic options to design prompts before constructing specific approaches. Furthermore, by focusing on zero-shot experiments, there will be a better understanding of PLMs and insights into novel ways of building language models.

The open question from our evaluation is that there are no convincing results with tweet characterization on a broad range of templates we experimented. So probably tweet characterization requires task-specific training. 

\bibliographystyle{ACM-Reference-Format}
\bibliography{etc}


\begin{thebibliography}{30}


\ifx \showCODEN    \undefined \def \showCODEN     #1{\unskip}     \fi
\ifx \showDOI      \undefined \def \showDOI       #1{#1}\fi
\ifx \showISBNx    \undefined \def \showISBNx     #1{\unskip}     \fi
\ifx \showISBNxiii \undefined \def \showISBNxiii  #1{\unskip}     \fi
\ifx \showISSN     \undefined \def \showISSN      #1{\unskip}     \fi
\ifx \showLCCN     \undefined \def \showLCCN      #1{\unskip}     \fi
\ifx \shownote     \undefined \def \shownote      #1{#1}          \fi
\ifx \showarticletitle \undefined \def \showarticletitle #1{#1}   \fi
\ifx \showURL      \undefined \def \showURL       {\relax}        \fi
\providecommand\bibfield[2]{#2}
\providecommand\bibinfo[2]{#2}
\providecommand\natexlab[1]{#1}
\providecommand\showeprint[2][]{arXiv:#2}

\bibitem[Barr{\'o}n-Cedeno et~al\mbox{.}(2019)]%
        {barron2019proppy}
\bibfield{author}{\bibinfo{person}{Alberto Barr{\'o}n-Cedeno},
  \bibinfo{person}{Israa Jaradat}, \bibinfo{person}{Giovanni Da~San~Martino},
  {and} \bibinfo{person}{Preslav Nakov}.} \bibinfo{year}{2019}\natexlab{}.
\newblock \showarticletitle{Proppy: Organizing the news based on their
  propagandistic content}.
\newblock \bibinfo{journal}{\emph{Information Processing \& Management}}
  \bibinfo{volume}{56}, \bibinfo{number}{5} (\bibinfo{year}{2019}),
  \bibinfo{pages}{1849--1864}.
\newblock


\bibitem[Bragg et~al\mbox{.}(2021)]%
        {bragg2021flex}
\bibfield{author}{\bibinfo{person}{Jonathan Bragg}, \bibinfo{person}{Arman
  Cohan}, \bibinfo{person}{Kyle Lo}, {and} \bibinfo{person}{Iz Beltagy}.}
  \bibinfo{year}{2021}\natexlab{}.
\newblock \showarticletitle{Flex: Unifying evaluation for few-shot nlp}.
\newblock \bibinfo{journal}{\emph{Advances in Neural Information Processing
  Systems}}  \bibinfo{volume}{34} (\bibinfo{year}{2021}).
\newblock


\bibitem[Clark et~al\mbox{.}(2019)]%
        {clark2019boolq}
\bibfield{author}{\bibinfo{person}{Christopher Clark}, \bibinfo{person}{Kenton
  Lee}, \bibinfo{person}{Ming-Wei Chang}, \bibinfo{person}{Tom Kwiatkowski},
  \bibinfo{person}{Michael Collins}, {and} \bibinfo{person}{Kristina
  Toutanova}.} \bibinfo{year}{2019}\natexlab{}.
\newblock \showarticletitle{BoolQ: Exploring the Surprising Difficulty of
  Natural Yes/No Questions}. In \bibinfo{booktitle}{\emph{Proceedings of the
  2019 Conference of the North American Chapter of the Association for
  Computational Linguistics: Human Language Technologies, Volume 1 (Long and
  Short Papers)}}. \bibinfo{pages}{2924--2936}.
\newblock


\bibitem[Davison et~al\mbox{.}(2019)]%
        {davison-etal-2019-commonsense}
\bibfield{author}{\bibinfo{person}{Joe Davison}, \bibinfo{person}{Joshua
  Feldman}, {and} \bibinfo{person}{Alexander Rush}.}
  \bibinfo{year}{2019}\natexlab{}.
\newblock \showarticletitle{Commonsense Knowledge Mining from Pretrained
  Models}. In \bibinfo{booktitle}{\emph{Proceedings of the 2019 Conference on
  Empirical Methods in Natural Language Processing and the 9th International
  Joint Conference on Natural Language Processing (EMNLP-IJCNLP)}}.
  \bibinfo{publisher}{Association for Computational Linguistics},
  \bibinfo{address}{Hong Kong, China}, \bibinfo{pages}{1173--1178}.
\newblock
\urldef\tempurl%
\url{https://doi.org/10.18653/v1/D19-1109}
\showDOI{\tempurl}


\bibitem[Gao et~al\mbox{.}(2021)]%
        {gao-etal-2021-making}
\bibfield{author}{\bibinfo{person}{Tianyu Gao}, \bibinfo{person}{Adam Fisch},
  {and} \bibinfo{person}{Danqi Chen}.} \bibinfo{year}{2021}\natexlab{}.
\newblock \showarticletitle{Making Pre-trained Language Models Better Few-shot
  Learners}. In \bibinfo{booktitle}{\emph{Proceedings of the 59th Annual
  Meeting of the Association for Computational Linguistics and the 11th
  International Joint Conference on Natural Language Processing (Volume 1: Long
  Papers)}}. \bibinfo{publisher}{Association for Computational Linguistics},
  \bibinfo{address}{Online}, \bibinfo{pages}{3816--3830}.
\newblock
\urldef\tempurl%
\url{https://doi.org/10.18653/v1/2021.acl-long.295}
\showDOI{\tempurl}


\bibitem[Goodman(1961)]%
        {goodman1961snowball}
\bibfield{author}{\bibinfo{person}{Leo~A Goodman}.}
  \bibinfo{year}{1961}\natexlab{}.
\newblock \showarticletitle{Snowball sampling}.
\newblock \bibinfo{journal}{\emph{The annals of mathematical statistics}}
  (\bibinfo{year}{1961}), \bibinfo{pages}{148--170}.
\newblock


\bibitem[Gururangan et~al\mbox{.}(2020)]%
        {gururangan2020don}
\bibfield{author}{\bibinfo{person}{Suchin Gururangan}, \bibinfo{person}{Ana
  Marasovi{\'c}}, \bibinfo{person}{Swabha Swayamdipta}, \bibinfo{person}{Kyle
  Lo}, \bibinfo{person}{Iz Beltagy}, \bibinfo{person}{Doug Downey}, {and}
  \bibinfo{person}{Noah~A Smith}.} \bibinfo{year}{2020}\natexlab{}.
\newblock \showarticletitle{Don’t Stop Pretraining: Adapt Language Models to
  Domains and Tasks}. In \bibinfo{booktitle}{\emph{Proceedings of the 58th
  Annual Meeting of the Association for Computational Linguistics}}.
  \bibinfo{pages}{8342--8360}.
\newblock


\bibitem[Hambardzumyan et~al\mbox{.}(2021)]%
        {hambardzumyan-etal-2021-warp}
\bibfield{author}{\bibinfo{person}{Karen Hambardzumyan}, \bibinfo{person}{Hrant
  Khachatrian}, {and} \bibinfo{person}{Jonathan May}.}
  \bibinfo{year}{2021}\natexlab{}.
\newblock \showarticletitle{{WARP}: {W}ord-level {A}dversarial
  {R}e{P}rogramming}. In \bibinfo{booktitle}{\emph{Proceedings of the 59th
  Annual Meeting of the Association for Computational Linguistics and the 11th
  International Joint Conference on Natural Language Processing (Volume 1: Long
  Papers)}}. \bibinfo{publisher}{Association for Computational Linguistics},
  \bibinfo{address}{Online}, \bibinfo{pages}{4921--4933}.
\newblock
\urldef\tempurl%
\url{https://doi.org/10.18653/v1/2021.acl-long.381}
\showDOI{\tempurl}


\bibitem[Hern(2018)]%
        {NewFaceb3:online}
\bibfield{author}{\bibinfo{person}{Alex Hern}.}
  \bibinfo{year}{2018}\natexlab{}.
\newblock \bibinfo{booktitle}{\emph{New Facebook controls aim to regulate
  political ads and fight fake news | Facebook | The Guardian}}.
\newblock
\urldef\tempurl%
\url{https://www.theguardian.com/technology/2018/apr/06/facebook-launches-controls-regulate-ads-publishers}
\showURL{%
\tempurl}
\newblock
\shownote{(Accessed on 01/04/2022)}.


\bibitem[Huffaker et~al\mbox{.}(2020)]%
        {huffaker2020crowdsourced}
\bibfield{author}{\bibinfo{person}{Jordan~S Huffaker},
  \bibinfo{person}{Jonathan~K Kummerfeld}, \bibinfo{person}{Walter~S Lasecki},
  {and} \bibinfo{person}{Mark~S Ackerman}.} \bibinfo{year}{2020}\natexlab{}.
\newblock \showarticletitle{Crowdsourced detection of emotionally manipulative
  language}. In \bibinfo{booktitle}{\emph{Proceedings of the 2020 CHI
  Conference on Human Factors in Computing Systems}}. \bibinfo{pages}{1--14}.
\newblock


\bibitem[Jiang et~al\mbox{.}(2020)]%
        {jiang2020x}
\bibfield{author}{\bibinfo{person}{Zhengbao Jiang}, \bibinfo{person}{Antonios
  Anastasopoulos}, \bibinfo{person}{Jun Araki}, \bibinfo{person}{Haibo Ding},
  {and} \bibinfo{person}{Graham Neubig}.} \bibinfo{year}{2020}\natexlab{}.
\newblock \showarticletitle{X-FACTR: Multilingual Factual Knowledge Retrieval
  from Pretrained Language Models}. In \bibinfo{booktitle}{\emph{Proceedings of
  the 2020 Conference on Empirical Methods in Natural Language Processing
  (EMNLP)}}. \bibinfo{pages}{5943--5959}.
\newblock


\bibitem[Kassner and Sch{\"u}tze(2020)]%
        {kassner2020negated}
\bibfield{author}{\bibinfo{person}{Nora Kassner} {and} \bibinfo{person}{Hinrich
  Sch{\"u}tze}.} \bibinfo{year}{2020}\natexlab{}.
\newblock \showarticletitle{Negated and Misprimed Probes for Pretrained
  Language Models: Birds Can Talk, But Cannot Fly}. In
  \bibinfo{booktitle}{\emph{Proceedings of the 58th Annual Meeting of the
  Association for Computational Linguistics}}. \bibinfo{pages}{7811--7818}.
\newblock


\bibitem[Khashabi et~al\mbox{.}(2018)]%
        {khashabi2018looking}
\bibfield{author}{\bibinfo{person}{Daniel Khashabi}, \bibinfo{person}{Snigdha
  Chaturvedi}, \bibinfo{person}{Michael Roth}, \bibinfo{person}{Shyam
  Upadhyay}, {and} \bibinfo{person}{Dan Roth}.}
  \bibinfo{year}{2018}\natexlab{}.
\newblock \showarticletitle{Looking beyond the surface: A challenge set for
  reading comprehension over multiple sentences}. In
  \bibinfo{booktitle}{\emph{Proceedings of the 2018 Conference of the North
  American Chapter of the Association for Computational Linguistics: Human
  Language Technologies, Volume 1 (Long Papers)}}. \bibinfo{pages}{252--262}.
\newblock


\bibitem[Kumar and Talukdar(2021)]%
        {DBLP:conf/acl/KumarT21}
\bibfield{author}{\bibinfo{person}{Sawan Kumar} {and}
  \bibinfo{person}{Partha~P. Talukdar}.} \bibinfo{year}{2021}\natexlab{}.
\newblock \showarticletitle{Reordering Examples Helps during Priming-based
  Few-Shot Learning}. In \bibinfo{booktitle}{\emph{Findings of the Association
  for Computational Linguistics: {ACL/IJCNLP} 2021, Online Event, August 1-6,
  2021}} \emph{(\bibinfo{series}{Findings of {ACL}},
  Vol.~\bibinfo{volume}{{ACL/IJCNLP} 2021})},
  \bibfield{editor}{\bibinfo{person}{Chengqing Zong}, \bibinfo{person}{Fei
  Xia}, \bibinfo{person}{Wenjie Li}, {and} \bibinfo{person}{Roberto Navigli}}
  (Eds.). \bibinfo{publisher}{Association for Computational Linguistics},
  \bibinfo{pages}{4507--4518}.
\newblock
\urldef\tempurl%
\url{https://doi.org/10.18653/v1/2021.findings-acl.395}
\showDOI{\tempurl}


\bibitem[Lester et~al\mbox{.}(2021)]%
        {lester2021power}
\bibfield{author}{\bibinfo{person}{Brian Lester}, \bibinfo{person}{Rami
  Al-Rfou}, {and} \bibinfo{person}{Noah Constant}.}
  \bibinfo{year}{2021}\natexlab{}.
\newblock \showarticletitle{The power of scale for parameter-efficient prompt
  tuning}.
\newblock \bibinfo{journal}{\emph{arXiv preprint arXiv:2104.08691}}
  (\bibinfo{year}{2021}).
\newblock


\bibitem[Li and Liang(2021)]%
        {li2021prefix}
\bibfield{author}{\bibinfo{person}{Xiang~Lisa Li} {and} \bibinfo{person}{Percy
  Liang}.} \bibinfo{year}{2021}\natexlab{}.
\newblock \showarticletitle{Prefix-tuning: Optimizing continuous prompts for
  generation}.
\newblock \bibinfo{journal}{\emph{arXiv preprint arXiv:2101.00190}}
  (\bibinfo{year}{2021}).
\newblock


\bibitem[Nishida et~al\mbox{.}(2020)]%
        {nishida2020unsupervised}
\bibfield{author}{\bibinfo{person}{Kosuke Nishida}, \bibinfo{person}{Kyosuke
  Nishida}, \bibinfo{person}{Itsumi Saito}, \bibinfo{person}{Hisako Asano},
  {and} \bibinfo{person}{Junji Tomita}.} \bibinfo{year}{2020}\natexlab{}.
\newblock \showarticletitle{Unsupervised Domain Adaptation of Language Models
  for Reading Comprehension}. In \bibinfo{booktitle}{\emph{Proceedings of the
  12th Language Resources and Evaluation Conference}}.
  \bibinfo{pages}{5392--5399}.
\newblock


\bibitem[Petroni et~al\mbox{.}(2019)]%
        {petroni2019language}
\bibfield{author}{\bibinfo{person}{Fabio Petroni}, \bibinfo{person}{Tim
  Rockt{\"a}schel}, \bibinfo{person}{Sebastian Riedel},
  \bibinfo{person}{Patrick Lewis}, \bibinfo{person}{Anton Bakhtin},
  \bibinfo{person}{Yuxiang Wu}, {and} \bibinfo{person}{Alexander Miller}.}
  \bibinfo{year}{2019}\natexlab{}.
\newblock \showarticletitle{Language Models as Knowledge Bases?}. In
  \bibinfo{booktitle}{\emph{Proceedings of the 2019 Conference on Empirical
  Methods in Natural Language Processing and the 9th International Joint
  Conference on Natural Language Processing (EMNLP-IJCNLP)}}.
  \bibinfo{pages}{2463--2473}.
\newblock


\bibitem[Radford et~al\mbox{.}(2019)]%
        {GPT2}
\bibfield{author}{\bibinfo{person}{Alec Radford}, \bibinfo{person}{Jeffrey Wu},
  \bibinfo{person}{Rewon Child}, \bibinfo{person}{David Luan},
  \bibinfo{person}{Dario Amodei}, \bibinfo{person}{Ilya Sutskever},
  {et~al\mbox{.}}} \bibinfo{year}{2019}\natexlab{}.
\newblock \showarticletitle{Language models are unsupervised multitask
  learners}.
\newblock \bibinfo{journal}{\emph{OpenAI blog}} \bibinfo{volume}{1},
  \bibinfo{number}{8} (\bibinfo{year}{2019}), \bibinfo{pages}{9}.
\newblock


\bibitem[Reynolds and McDonell(2021)]%
        {reynolds2021prompt}
\bibfield{author}{\bibinfo{person}{Laria Reynolds} {and} \bibinfo{person}{Kyle
  McDonell}.} \bibinfo{year}{2021}\natexlab{}.
\newblock \showarticletitle{Prompt programming for large language models:
  Beyond the few-shot paradigm}. In \bibinfo{booktitle}{\emph{Extended
  Abstracts of the 2021 CHI Conference on Human Factors in Computing Systems}}.
  \bibinfo{pages}{1--7}.
\newblock


\bibitem[Schick et~al\mbox{.}(2020)]%
        {schick2020automatically}
\bibfield{author}{\bibinfo{person}{Timo Schick}, \bibinfo{person}{Helmut
  Schmid}, {and} \bibinfo{person}{Hinrich Sch{\"u}tze}.}
  \bibinfo{year}{2020}\natexlab{}.
\newblock \showarticletitle{Automatically Identifying Words That Can Serve as
  Labels for Few-Shot Text Classification}. In
  \bibinfo{booktitle}{\emph{Proceedings of the 28th International Conference on
  Computational Linguistics}}. \bibinfo{pages}{5569--5578}.
\newblock


\bibitem[Schick and Sch{\"u}tze(2021a)]%
        {schick2021exploiting}
\bibfield{author}{\bibinfo{person}{Timo Schick} {and} \bibinfo{person}{Hinrich
  Sch{\"u}tze}.} \bibinfo{year}{2021}\natexlab{a}.
\newblock \showarticletitle{Exploiting Cloze-Questions for Few-Shot Text
  Classification and Natural Language Inference}. In
  \bibinfo{booktitle}{\emph{Proceedings of the 16th Conference of the European
  Chapter of the Association for Computational Linguistics: Main Volume}}.
  \bibinfo{pages}{255--269}.
\newblock


\bibitem[Schick and Sch{\"u}tze(2021b)]%
        {schick2021s}
\bibfield{author}{\bibinfo{person}{Timo Schick} {and} \bibinfo{person}{Hinrich
  Sch{\"u}tze}.} \bibinfo{year}{2021}\natexlab{b}.
\newblock \showarticletitle{It’s Not Just Size That Matters: Small Language
  Models Are Also Few-Shot Learners}. In \bibinfo{booktitle}{\emph{Proceedings
  of the 2021 Conference of the North American Chapter of the Association for
  Computational Linguistics: Human Language Technologies}}.
  \bibinfo{pages}{2339--2352}.
\newblock


\bibitem[Sherr(2018)]%
        {Facebook61:online}
\bibfield{author}{\bibinfo{person}{Ian Sherr}.}
  \bibinfo{year}{2018}\natexlab{}.
\newblock \bibinfo{booktitle}{\emph{Facebook and Twitter have new rules for
  political ads. Here's how they work}}.
\newblock
\urldef\tempurl%
\url{https://www.cnet.com/news/facebook-and-twitter-have-new-rules-for-political-ads-heres-how-they-work/}
\showURL{%
\tempurl}
\newblock
\shownote{(Accessed on 01/04/2022)}.


\bibitem[Spinde et~al\mbox{.}(2020)]%
        {spinde2020enabling}
\bibfield{author}{\bibinfo{person}{Timo Spinde}, \bibinfo{person}{Felix
  Hamborg}, \bibinfo{person}{Karsten Donnay}, \bibinfo{person}{Angelica
  Becerra}, {and} \bibinfo{person}{Bela Gipp}.}
  \bibinfo{year}{2020}\natexlab{}.
\newblock \showarticletitle{Enabling news consumers to view and understand
  biased news coverage: a study on the perception and visualization of media
  bias}. In \bibinfo{booktitle}{\emph{Proceedings of the ACM/IEEE joint
  conference on digital libraries in 2020}}. \bibinfo{pages}{389--392}.
\newblock


\bibitem[Wang et~al\mbox{.}(2019)]%
        {wang2019superglue}
\bibfield{author}{\bibinfo{person}{Alex Wang}, \bibinfo{person}{Yada
  Pruksachatkun}, \bibinfo{person}{Nikita Nangia}, \bibinfo{person}{Amanpreet
  Singh}, \bibinfo{person}{Julian Michael}, \bibinfo{person}{Felix Hill},
  \bibinfo{person}{Omer Levy}, {and} \bibinfo{person}{Samuel~R Bowman}.}
  \bibinfo{year}{2019}\natexlab{}.
\newblock \showarticletitle{SuperGLUE: a stickier benchmark for general-purpose
  language understanding systems}. In \bibinfo{booktitle}{\emph{Proceedings of
  the 33rd International Conference on Neural Information Processing Systems}}.
  \bibinfo{pages}{3266--3280}.
\newblock


\bibitem[Wang et~al\mbox{.}(2018)]%
        {wang2018glue}
\bibfield{author}{\bibinfo{person}{Alex Wang}, \bibinfo{person}{Amanpreet
  Singh}, \bibinfo{person}{Julian Michael}, \bibinfo{person}{Felix Hill},
  \bibinfo{person}{Omer Levy}, {and} \bibinfo{person}{Samuel Bowman}.}
  \bibinfo{year}{2018}\natexlab{}.
\newblock \showarticletitle{GLUE: A Multi-Task Benchmark and Analysis Platform
  for Natural Language Understanding}. In \bibinfo{booktitle}{\emph{Proceedings
  of the 2018 EMNLP Workshop BlackboxNLP: Analyzing and Interpreting Neural
  Networks for NLP}}. \bibinfo{pages}{353--355}.
\newblock


\bibitem[Wei et~al\mbox{.}(2020)]%
        {wei2020examining}
\bibfield{author}{\bibinfo{person}{Kai Wei}, \bibinfo{person}{Yu-Ru Lin}, {and}
  \bibinfo{person}{Muheng Yan}.} \bibinfo{year}{2020}\natexlab{}.
\newblock \showarticletitle{Examining Protest as An Intervention to Reduce
  Online Prejudice: A Case Study of Prejudice Against Immigrants}. In
  \bibinfo{booktitle}{\emph{Proceedings of The Web Conference 2020}}.
  \bibinfo{pages}{2443--2454}.
\newblock


\bibitem[YouTube(2018)]%
        {Building9:online}
\bibfield{author}{\bibinfo{person}{YouTube}.} \bibinfo{year}{2018}\natexlab{}.
\newblock \bibinfo{booktitle}{\emph{Building a better news experience on
  YouTube, together}}.
\newblock
\urldef\tempurl%
\url{https://blog.youtube/news-and-events/building-better-news-experience-on/}
\showURL{%
\tempurl}
\newblock
\shownote{(Accessed on 01/04/2022)}.


\bibitem[Zhang et~al\mbox{.}(2018)]%
        {zhang2018record}
\bibfield{author}{\bibinfo{person}{Sheng Zhang}, \bibinfo{person}{Xiaodong
  Liu}, \bibinfo{person}{Jingjing Liu}, \bibinfo{person}{Jianfeng Gao},
  \bibinfo{person}{Kevin Duh}, {and} \bibinfo{person}{Benjamin Van~Durme}.}
  \bibinfo{year}{2018}\natexlab{}.
\newblock \showarticletitle{Record: Bridging the gap between human and machine
  commonsense reading comprehension}.
\newblock \bibinfo{journal}{\emph{arXiv preprint arXiv:1810.12885}}
  (\bibinfo{year}{2018}).
\newblock


\end{thebibliography}

\appendix

\end{document}